RESEARCH

# Enhancing deep learning models for time series classification via knowledge distillation

**Javidan Abdullayev[1] · Maxime Devanne[1] · Jonathan Weber[1] · Germain Forestier[1,2]**

## Abstract

Deep learning has achieved remarkable success in various domains including time series analysis, computer vision and natural language processing. However, high computational and memory demands of state-of-the-art architectures pose challenges for deployment in resource-limited environments. Knowledge Distillation (KD) addresses this by transferring knowledge from a large teacher model to a smaller, more efficient student model while maintaining competitive performance. In this work, we investigate the effectiveness of KD for Time Series Classification (TSC) across three architectures: the classical *Fully Convolutional Network* (FCN), the convolutional *Inception* model and the transformer-based *ConvTran* model. We evaluate our approach on UCR Archive, the largest benchmark repository of time series datasets, by modifying architectural components such as convolutional filters, Inception modules and attention heads across the three architectures. Our results consistently show that KD most effectively benefits student models of intermediate complexity across all three architectures, with the distilled FCN student reducing parameters by a factor of 38, the distilled Inception student achieving nearly the same performance as the teacher with 42% fewer parameters and the distilled ConvTran student with 2 attention heads showing the most significant improvement through distillation. To encourage further research and reproducibility, we provide our implementation at https://github.com/MSD-IRIMAS/KD-4-TSC.

Maxime Devanne, Jonathan Weber and Germain Forestier have contributed equally to this work.

✉ Javidan Abdullayev
javidan.abdullayev@uha.fr

Maxime Devanne
maxime.devanne@uha.fr

Jonathan Weber
jonathan.weber@uha.fr

Germain Forestier
germain.forestier@uha.fr

[1] IRIMAS, University of Haute Alsace, Frères Lumière, Mulhouse 68200, France

[2] Faculty of IT, Monash University, Wellington Road, Melbourne 3800, Australia

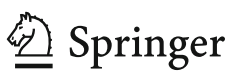

## 1 Introduction

Deep learning has fundamentally transformed time series analysis, enabling automatic feature extraction directly from raw data without the need for manual feature engineering [20]. A significant amount of time series data has recently been exploited for various applications such as human motion [11], satellite imagery [8] and healthcare [33]. The growing volume of time series data and having the same topology as image data with one less dimension [5] led researchers to investigate deep learning approaches in time series analysis. Recently, time series classification (TSC) has emerged as one of the most active research topics with applications across a wide range of fields [28, 50]. Many of the well-performed TSC methods excluding those based on deep learning involve feature engineering as a separate task prior to classification. Those approaches normally result in information loss and complicate the development process. In contrast, deep learning models perform feature engineering automatically as part of the learning process and remove the need for manual feature engineering. Consequently, they can extract information faster, more directly and completely from the given time series data. In this work, we focus on the task of TSC, exploring deep neural network architectures ranging from convolutional to transformer-based models [12, 20].

Empirically, neural networks tend to perform better as their complexity grows which has led to increasingly complex state-of-the-art deep learning models for TSC. One significant drawback of such complex DNNs is the difficulty of deploying them in real-world environments. Deploying very deep neural networks in real-time applications or on embedded devices with limited resources is challenging due to their substantial computational and memory requirements.

To enhance efficiency while maintaining high performance, recent works have focused on knowledge distillation (KD), one of the most prominent model compression techniques. KD involves transferring knowledge from a deep pre-trained model to a shallower model which aims to replicate the outputs of the larger model. In this context, a deeper model is commonly referred to as a *teacher* model and a shallower model as a *student* model. [7] initially proposed this concept to transfer knowledge from a larger network to a smaller network without causing a significant loss in performance. Later [17] proposed a variation in this method which became known as *knowledge distillation*. Findings in these papers convincingly demonstrate that a shallower model can achieve competitive or superior performance in comparison with a very deep model.

With the development of deeper neural network architectures, the success of knowledge distillation has been validated across various fields of artificial intelligence, including computer vision [14], text recognition [42] and natural language processing [15]. Only a small number of papers have investigated knowledge distillation in the field of TSC. [1] explored the impact of knowledge distillation using the classical FCN architecture for TSC. Another work [13] improved the performance of a *student* model by rectifying the incorrectly predicted soft labels from the teacher Inception model. Recent advances in large-scale knowledge distillation have further demonstrated its potential in improving model efficiency while retaining high performance, most notably in transformer-based models such as DeepSeek [16].

In our previous study [1], we investigated knowledge distillation exclusively for the FCN architecture, focusing on reducing its convolutional filters. In this paper, we significantly extend our study along two directions. First, we incorporate two additional architectures, namely the convolutional Inception model [21] and the transformer-based ConvTran model [12], analyzing effects of KD on Inception modules and attention heads, respectively. Second, we provide a deeper analysis of the distillation process by examining

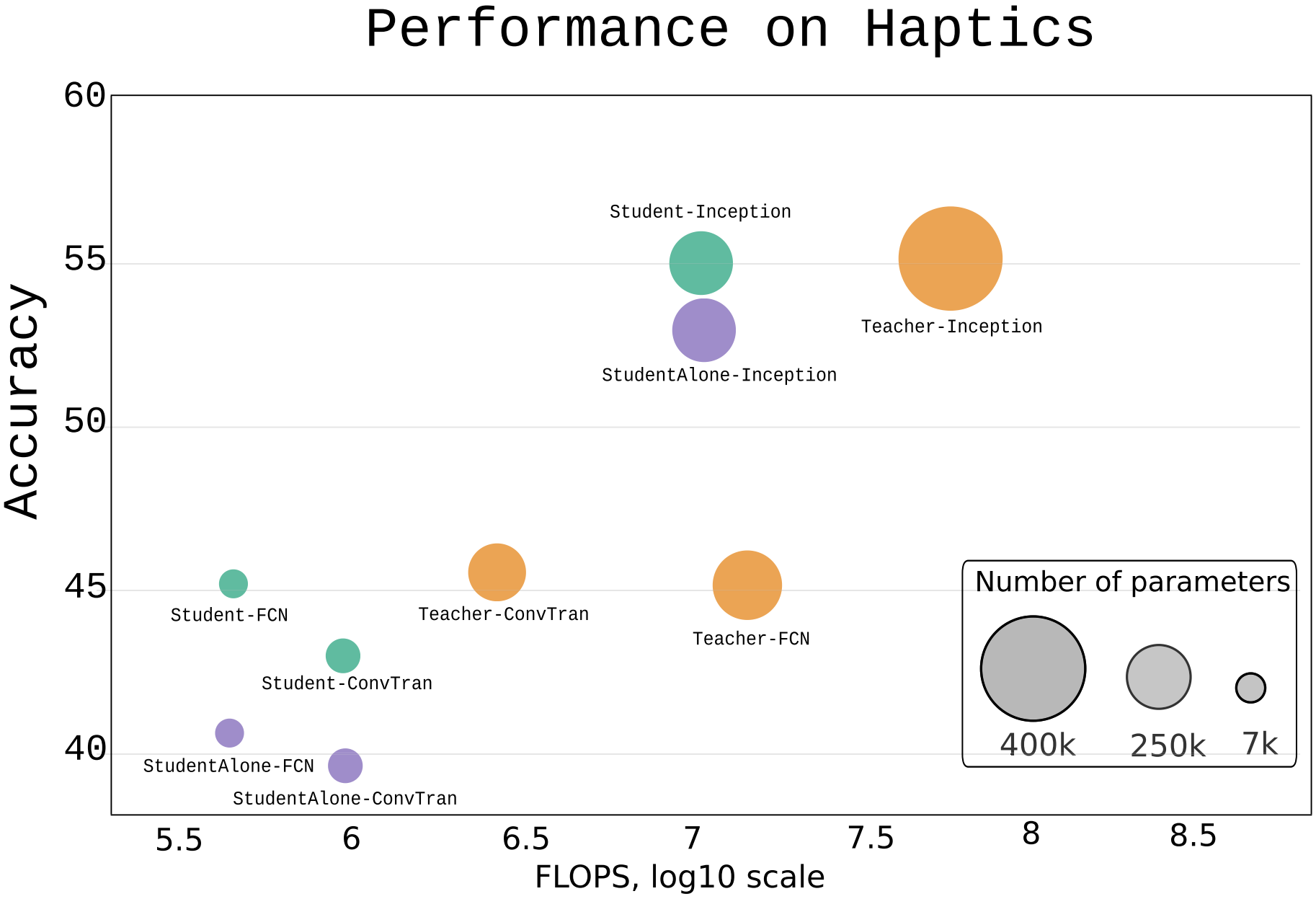

**Fig. 1** Accuracy plot shows performances of the *student*, *studentAlone (vanilla student)* and *teacher* models for FCN, Inception and ConvTran networks on the Haptics dataset

filter space transfer between teacher and student models and investigating how KD affects generalization behavior.

Figure 1 depicts an accuracy plot for the performances of *teacher*, *student* and *studentAlone (vanilla student)* models in the case of FCN [41], Inception [21] and ConvTran [12] architectures on the Haptics dataset. The *studentAlone* model represents the same network as the *student* model trained without knowledge distillation. The Haptics dataset from the UCR Archive contains time series data representing tactile sensor responses, used for classifying different types of interactions in haptic-based systems. The use of knowledge distillation for TSC is evaluated on the UCR Archive 2018 [9], the largest available repository of time series datasets. Experiments show that knowledge distillation is especially beneficial for *student* models of intermediate complexity.

The remainder of this paper is arranged as follows: The related work is introduced in Sect. 2. Next, we provide detailed background information about TSC, knowledge distillation and teacher–student model configurations in Sect. 3. All experiments and results are discussed in Sect. 4. Finally, the conclusion is given in Sect. 5.

## 2 Related work

### 2.1 Knowledge distillation

Knowledge distillation (KD) is a technique to transfer knowledge from a deep model or an ensemble of models to a smaller, shallower model. This concept was initially introduced by [7]. They suggested that minimizing the divergence between the output logits of *teacher* and *student* models allows the *student* to mimic the *teacher* predictions. Consequently, the distilled *student* network achieves competitive performance despite having fewer parameters.

Building on this concept, [17] proposed training the student model using output of the Softmax function so that the Softmax outputs of the *student* and *teacher* models are similar. They also introduced a temperature parameter to soften probability distributions which helps the *teacher* model transfer its generalization capabilities to the *student* model more effectively. This technique was further extended in [34] by suggesting the use of feature maps of intermediate layers as hints to enhance the performance of the student model. Similarly, [48] considered intermediate representations while [31] incorporated metric learning into knowledge distillation through a triplet loss function. An alternative approach, the teacher assistant method, was introduced in [29] where a model of intermediate complexity bridges the gap between a shallower *student* and a deeper *teacher* model.

Although knowledge distillation has been applied primarily in computer vision [14, 44], its promising results have inspired research in various other fields, including document analysis [36], speech processing [38] and natural language processing [37]. In the domain of time series analysis, a small number of studies have explored the use of knowledge distillation for tasks such as regression [43], forecasting [18] and classification [1, 13]. More recently, knowledge distillation has been scaled to large language models to improve efficiency and scalability [16, 45].

We note that recent research has proposed lightweight architectures for *time series forecasting* such as DLinear [49] and FITS [46] which reduce model complexity using linear modeling and frequency-domain representations. However, these forecasting-oriented models fall outside the scope of this study. In this work, we investigate the adaptation of knowledge distillation for *time series classification*, focusing on its effectiveness in reducing model complexity while maintaining performance across three architectures: classical FCN, convolutional Inception model and transformer-based ConvTran model.

## 2.2 Time series classification

Time series classification is one of the most challenging research areas in data mining [47]. Due to the significant increase in the amount of available time series data, many TSC methods have been proposed in recent years [3]. One of the well-known methods for TSC is to employ 1-Nearest Neighbor (NN) with a distance function [26] which has been the subject of many studies. Among distance functions, Dynamic Time Warping (DTW) has been widely considered as a strong baseline for NN-based TSC, demonstrating promising results across a wide range of datasets [3]. Building on DTW, [32] proposed a novel DTW averaging algorithm that improves both classification accuracy and computational efficiency. To further enhance performance, some studies have proposed DTW variants that incorporate temporal constraints such as Time-weight-based DTW (TDTW) [24] and Adaptive constrained DTW (ACDTW) [25]. Other research efforts have focused on enhancing time series data representation with notable approaches including interpretable representations [4] and Shape-sphere [23]. A comparative study [26] evaluated several other distance functions combined with NN, showing that none of them significantly outperformed DTW. The same study also showed that ensembling individual NN classifiers with different distance measures outperforms every single component of the ensemble. To surpass the NN-DTW classifier, subsequent works focused on developing ensembles of different classifiers built on multiple time series representations [2]. These representations can be constructed in various ways, with the most notable techniques being shapelet transformers [6], DTW features [22] and time series feature learning [40].

Building on this idea, researchers proposed COTE (Collective Of Transformation-based Ensembles) [2], an ensemble of 35 classifiers built on multiple time series representations. COTE was later extended using a Hierarchical Vote system, resulting in HIVE-COTE 2.0 [27], one of the strongest performing approaches in TSC. However, these ensemble models are computationally expensive which makes them challenging to deploy in resource-limited environments.

In addition to these approaches, dictionary-based methods have also been explored for TSC. A notable example is BOSS (Bag-of-SFA Symbols), introduced by [35], which uses symbolic approximations of time series combined with an ensemble classifier to achieve high accuracy, even in the presence of noise. Another work [30] introduced closed motifs for streaming time series classification, providing a new perspective on the efficient handling of continuous, real-time data.

The rapid development of deep learning has also had a significant impact on TSC with numerous deep learning-based architectures proposed in recent years. A comprehensive review by [20] compared several deep learning-based TSC architectures which provide valuable insights into their relative strengths and limitations. The results showed that convolution-based architectures, such as FCNs and ResNets, achieved state-of-the-art performance on UCR Archive datasets. InceptionTime [21] adapts the Inception architecture for TSC by employing convolutional filters of various sizes, demonstrating that multi-scale feature extraction leads to improved classification performance. ROCKET [10] addresses the long training time of deep learning-based approaches by using a large number of random convolution kernels as feature extractors combined with a Ridge regression classifier which achieves competitive performance with significantly reduced training time. A recent approach, LITETime [19], which uses depth-wise separable convolutions with boosting techniques, demonstrated similar performance to the state-of-the-art InceptionTime model while requiring 2.78 times less training time. Meaningful representations of time series data in latent spaces have also been explored through multitasking neural networks [39], which leads to better estimation of time series averages and improved classification performance. In addition, attention-based architectures have gained increasing interest in TSC, with ConvTran [12] being a notable example which combines convolutional and self-attention mechanisms to capture both local and global temporal dependencies.

In this paper, our aim is not to develop a state-of-the-art model but rather to investigate the effectiveness of knowledge distillation for TSC across convolutional and transformer-based architectures and analyze the obtained results from different perspectives including model compression, filter space transfer and generalization behavior.

## 3 Knowledge distillation for time series classification

### 3.1 Background on deep learning for TSC

A time series is a sequence of data points that represents the evolution of a specific quantity over equal time increments. Let $t_1, t_2, ..., t_T$ be timestamps then a time series can be represented as a vector in this form $\mathbf{X} = [x_{t_1}, x_{t_2}, x_{t_3}, ..., x_{t_T}]$, $x \in R^T$, where each element in the vector represents the value of the quantity at a specific timestamp. In the case of univariate time series, each element is a scalar value, $x_{t_i} \in R, i = 1, ..., T$. In contrast, a multivariate time series consists of multiple variables observed simultaneously at each timestamp where each observation is a vector $x_{t_i} \in R^M$ with $M$ denoting the number of variables. In this work, we focus on univariate time series classification.

The true label or hard label for a given time series example is defined as a one-hot vector $Y = \{y_1, y_2, ..., y_C\}$ where C represents the total number of classes. The soft label or soft target for a given time series example is defined as a vector $\hat{Y} = \{\hat{y_1}, \hat{y_2}, ..., \hat{y_C}\}$ where $\hat{Y} \in [0, 1]$ and indicates the probability that x belongs to class $c \in \{1, 2, ..., C\}$. A dataset $\mathcal{D}$ is a collection of pairs $\{(X_1, Y_1), (X_2, Y_2), ..., (X_n, Y_n)\}$ where $X_i$ is a univariate time series and $Y_i$ is its corresponding one-hot label vector.

In TSC, the primary task is to assign a correct class label to a given time series example. Therefore, the objective is to estimate a function $\mathcal{F}$ to map from a given input space $X$ to an output space $Y$. In the context of deep learning, such a function corresponds to a Deep Neural Network (DNN) model. Since true labels are available during the training of deep learning models, the goal is to optimize the model parameters so that predicted labels $\hat{Y}$ match true labels $Y$. To evaluate performance of the mapping function, a loss function $\mathcal{L}$ measures the difference between the predicted and true labels. The categorical cross-entropy loss function $\mathcal{L}_{\mathcal{CE}}$ is commonly used in deep learning-based TSC tasks.

As shown in [20], deep learning architectures that achieve outstanding performance in TSC include Convolutional Neural Networks (CNNs) and more recently, transformer-based models [12]. A CNN is a type of neural network architecture that contains convolutional layers to compute 1D temporal convolutions on given time series. A convolution operation involves sliding a filter $\omega$ of size $l$ over a time series $X$ of size $T$. For an arbitrary timestamp $t$, a convolution operation $C_t$ in a convolution layer can be defined as

$$C_t = g(\omega * X_{t-l/2:t+l/2} + b) \mid \forall t \in [1, T], \tag{1}$$

where $g$ is the activation function which is typically a nonlinear function, $b$ is the bias term and $*$ denotes the convolution process.

In addition to convolutional architectures, transformer-based models have also been applied to TSC. These models rely on self-attention mechanisms to capture global dependencies across the entire time series, complementing the local feature extraction capability of convolutional operations.

## 3.2 Knowledge distillation for TSC

KD is a collaborative model compression technique which helps to reduce the complexity of neural network architectures while preserving performance. The complexity of a neural network is determined by its architectural components such as the number of layers, filters and neurons. For transformer-based models, additional factors include the number of attention heads and the embedding dimension. The size of a neural network is generally considered to be the number of parameters it contains. KD involves two neural network architectures with different complexities. One is a larger, pre-trained network known as the *teacher* model and the other one is a smaller network, the *student* model. Typically, the larger network (*teacher*) has more ability to fit well with a given dataset and consequently performs better than the smaller network (*student*). The essence of knowledge distillation is the transfer of knowledge from the *teacher* model to the *student* model during the training of the *student* model. This collaborative process allows the *teacher* model to guide the training of the *student* model, improving its performance. As a result, the distilled *student* model is expected to surpass the *studentAlone* (vanilla student) model and have competitive performance compared to the *teacher* model.

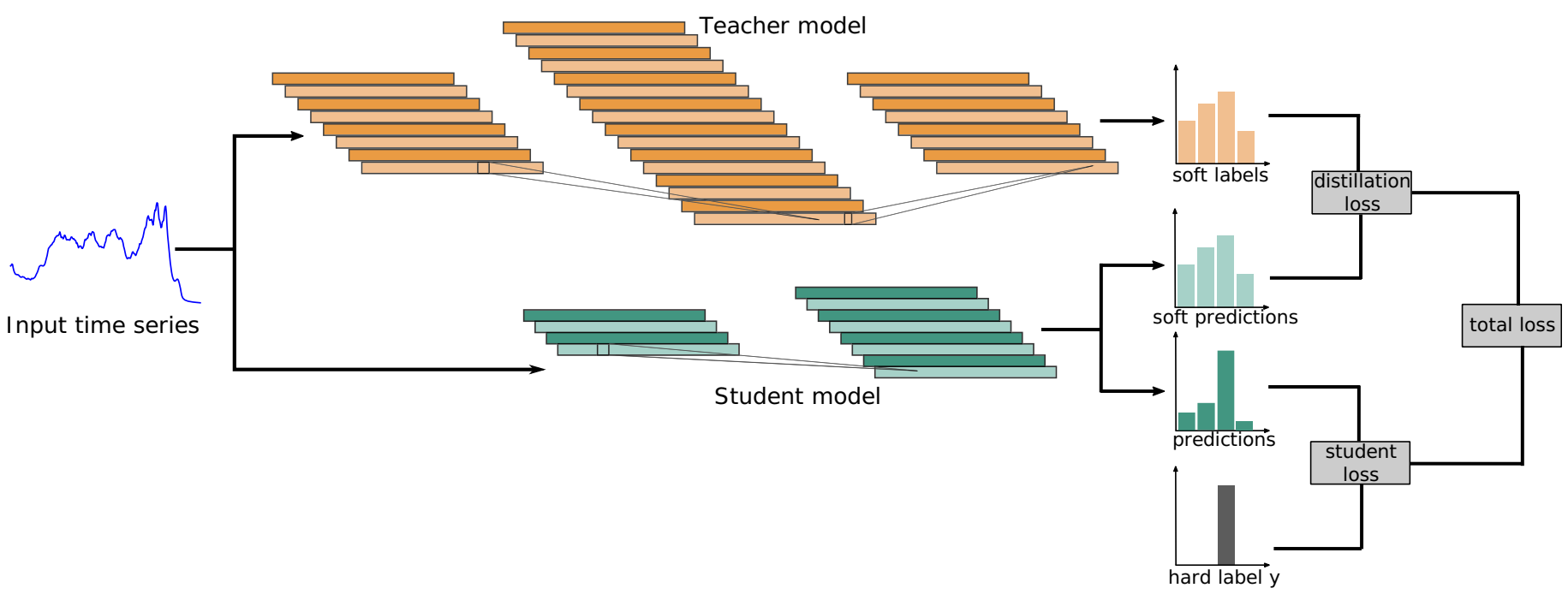

**Fig. 2** Teacher–Student FCN architecture for knowledge distillation

In the context of TSC, a *student* model is trained using both the dataset D and the predictions of a *teacher* model. Figure 2 illustrates the general architecture of the knowledge distillation process applied to the FCN network [41]. The figure demonstrates how the *student* model learns to replicate the predictions of the *teacher* model. This approach aims to minimize the disparity between the outputs of the *student* and *teacher* models, enhancing the *student* model performance by leveraging the *teacher* model knowledge.

As illustrated in Fig. 2, an input time series is simultaneously fed to both the *teacher* and *student* models, propagating through each network. Subsequently, the distillation loss and student loss are calculated separately. As a student loss, we employ the cross-entropy function, the most frequently used loss function in classification problems, as defined in equation 2.

$$\mathcal{L}_{CE}(\mathbf{Y}, \hat{\mathbf{Y}}_S) = -\sum_{i=1}^{C} \mathbf{Y}_i \log(\hat{\mathbf{Y}}_{S_i}) \tag{2}$$

where $Y$ denotes true labels and $\hat{Y_S}$ denotes the predicted class probabilities of the *student* model corresponding to the input. The predicted values $\hat{Y_S}$ are actual class probabilities computed by applying the Softmax function to the output logits of the last layer of the *student* model. The formula for the Softmax function is given in Eq. 3,

$$\hat{\mathbf{Y}}_i = Softmax(z_i) = \frac{e^{z_i}}{\sum_{j=1}^{C} e^{z_j}} \tag{3}$$

where $\mathbf{z}$ denotes the logit vector from the last layer.

Next, the Kullback–Leibler (KL) divergence function is typically used to calculate the distillation loss. This method measures the diversity between two probability distributions. In the context of KD, these probability distributions are the softened predicted probabilities of the *teacher* and *student* models. The formula to calculate KL loss is given in Eq. 4.

$$\mathcal{L}_{KL}(\hat{\mathbf{Y}}_T^{\tau}, \hat{\mathbf{Y}}_S^{\tau}) = -\sum_{i=1}^{C} \hat{\mathbf{Y}}_{T_i}^{\tau} \log \frac{\hat{\mathbf{Y}}_{T_i}^{\tau}}{\hat{\mathbf{Y}}_{S_i}^{\tau}} \tag{4}$$

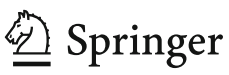

where $\hat{Y_T}^\tau$ and $\hat{Y_S}^\tau$ denote softened predicted probabilities of the *teacher* and *student* models, respectively.

The softened predicted probabilities are obtained by applying the Softmax function with a temperature parameter $\tau$ to the output logits of *teacher* and *student* models. The Softmax function with a temperature parameter $\tau$ is depicted in Eq. 5

$$\hat{\mathbf{Y}}_i^\tau = Softmax(z_i/\tau) = \frac{e^{z_i/\tau}}{\sum_{j=1}^{C} e^{z_j/\tau}} \tag{5}$$

where $\mathbf{z}$ represents the logit vector from the last layer and $\tau$ represents a temperature parameter.

As shown in Eq. 5, when $\tau$ is equal to one, it calculates the same values as the original Softmax function 3. As the value of $\tau$ increases, the predicted probability distributions for both models become smoother. The primary reason for applying the Softmax function with a temperature parameter $\tau$ is to soften the probability distributions, facilitating knowledge transfer from the *teacher* to the *student* model.

The knowledge acquired from a *teacher* model during the training of a *student* model is captured in the distillation loss. The total loss is calculated as a weighted sum of the student loss and the distillation loss. The backpropagation algorithm then optimizes the parameters of the *student* network based on the total loss. The formula for calculating the total loss is given in Eq. 6.

$$L_{KD} = \lambda \times L_{CE}(\mathbf{Y}, \hat{\mathbf{Y}}_S) + (1 - \lambda) \times L_{KL}(\hat{\mathbf{Y}}_T^\tau, \hat{\mathbf{Y}}_S^\tau) \tag{6}$$

where $Y$, $\hat{Y}_S$, $\hat{Y}_T^\tau$ and $\hat{Y}_S^\tau$ represent hard predictions, standard *student* predictions (output of Softmax function), softened *teacher* predictions with a temperature value $\tau$ and softened *student* predictions with the same temperature value $\tau$, respectively. The parameter $\lambda$ is used to adjust the relative impact of the student and distillation losses in the total loss function.

### 3.3 Teacher models

As previously mentioned, deep learning architectures including convolutional and transformer-based models have demonstrated state-of-the-art results for Time Series Classification (TSC) [12, 20]. For our experiments, we selected three such architectures: FCN [41], Inception [21] and ConvTran [12]. The rationale for choosing these three architectures is to evaluate the effectiveness of knowledge distillation across models of varying complexity and design paradigms, from a relatively lightweight convolutional architecture (FCN) to a more complex convolutional model (Inception) and a transformer-based model (ConvTran). Our primary objective is to analyze the impact of knowledge distillation by reducing the number of convolutional filters in the FCN architecture, the number of Inception modules in the Inception architecture and the number of attention heads and embedding dimensions in the ConvTran architecture.

#### 3.3.1 FCN network

The *teacher* FCN model, first proposed in [41], consists of three convolutional blocks. Each block includes three operations: 1D convolution, batch normalization and ReLU activation. An input time series first undergoes a 1D convolution operation, followed by batch normalization and then passes through a ReLU activation. In all convolutional blocks, 1D convolutions have a stride of 1 and zero padding to preserve the length of the input time series after the convolution operation. The first convolutional block has 128 filters of size 8, followed by a second block with 256 filters of size 5 and the third block contains 128 filters of size 3. The output of the last convolutional block is averaged over the temporal dimension using a Global Average Pooling (GAP) layer and the result is then fed into the Softmax classifier. The teacher FCN architecture comprises 267,018 model parameters as shown in Fig. 2 and summarized in Table 1.

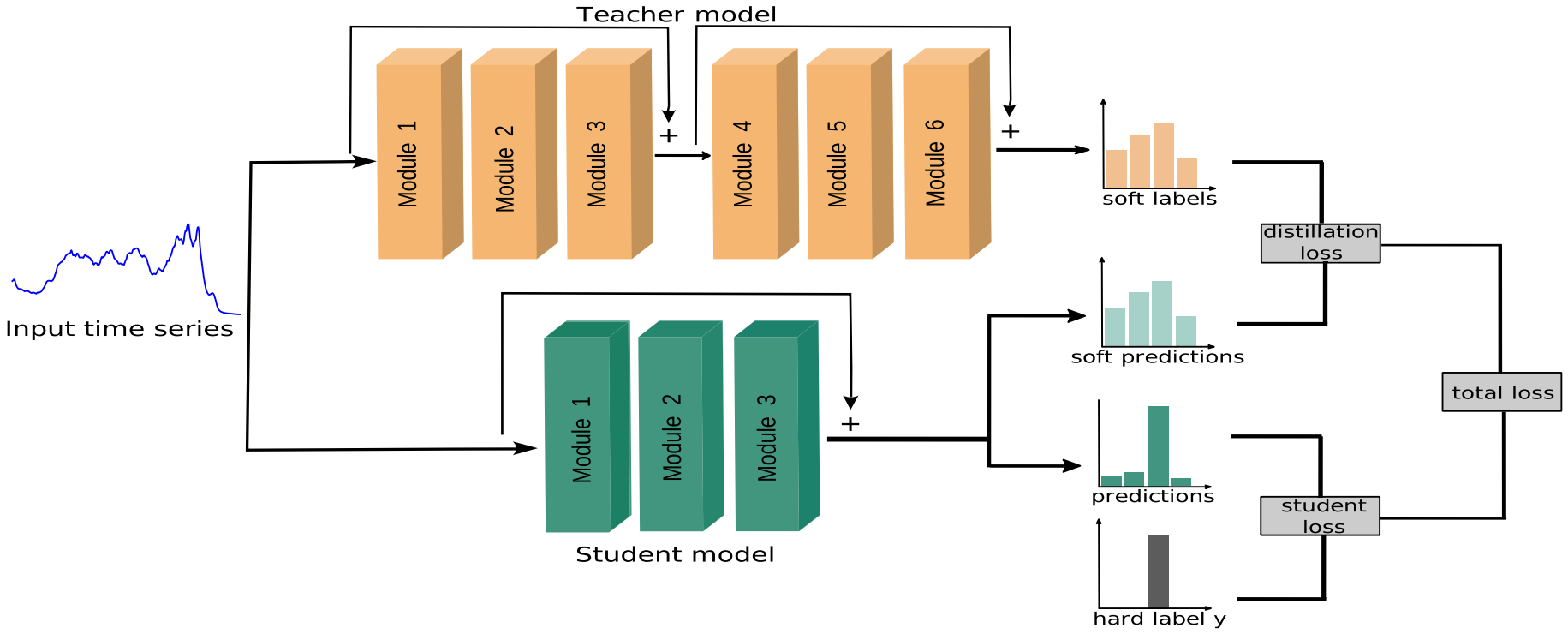

**Fig. 3** Teacher–Student inception architecture for knowledge distillation

### 3.3.2 Inception network

The *teacher* Inception model was first introduced in [21], and its overall architecture is depicted in Fig. 3. As shown in the figure, the *teacher* Inception network includes two residual blocks. Each residual block comprises three Inception modules. Each Inception module has three convolutional branches with kernel sizes of 10, 20 and 40, along with a pooling branch to capture features at various scales and resolutions. The output of an Inception module is created by concatenating the outputs of all its branches along the depth dimension. The input to each residual block is added to its output using a skip linear connection. After the final Inception module, a Global Average Pooling (GAP) layer is applied to reduce the spatial dimensions of the output multivariate time series. Finally, a Softmax activation function is applied to transform logits of the model into a class probability distribution. The architecture of the *teacher* Inception model includes 420,708 model parameters, as summarized in Table 2.

### 3.3.3 ConvTran network

The *teacher* ConvTran model, introduced in [12], combines convolutional and self-attention mechanisms for TSC. The detailed architecture of ConvTran is described in [12]. The original ConvTran architecture was designed for multivariate time series classification where the embedding layer consists of two convolutional operations: a temporal convolution to capture local temporal patterns and a spatial convolution to capture dependencies among channels. Since we evaluate on univariate time series datasets from the UCR Archive, we remove the spatial convolution from the embedding layer and retain only the temporal convolution. The output of the embedding layer is then passed through a transformer encoder that uses multi-head self-attention to capture global temporal dependencies. The *teacher* ConvTran model uses 8 attention heads with an embedding dimension of 16, comprising 33,026 model parameters as summarized in Table 3.

**Table 1** Configurations of *teacher* FCN and *student* FCN architectures

| | Model name | # filters layer 1 | # filters layer 2 | # filters layer 3 | Total # parameters |
|---|---|---|---|---|---|
| Teacher | Model_F128 | 128 | 256 | 128 | 267 018 |
| Students | Model_64F | 64 | 128 | 64 | 67 978 |
| | Model_F32 | 32 | 64 | 32 | 17 610 |
| | Model_F28 | 28 | 56 | 28 | 13 618 |
| | Model_F24 | 24 | 48 | 24 | 10 138 |
| | Model_F20 | 20 | 40 | 20 | 7 170 |
| | Model_F16 | 16 | 32 | 16 | 4 714 |
| | Model_F12 | 12 | 24 | 12 | 2 770 |
| | Model_F8 | 8 | 16 | 8 | 1 338 |
| | Model_F4 | 4 | 8 | 4 | 418 |

## 3.4 Student models

To achieve successful knowledge distillation, designing appropriate *student* architectures is crucial. The literature presents various methods for creating suitable *student* models for a specific *teacher* network. For effective knowledge transfer, it is always recommended to make the *student* architecture similar to the *teacher* architecture with fewer layers or fewer filters in each layer [14]. As discussed earlier, our goal is to evaluate the performance of knowledge distillation by reducing the number of convolutional filters in the FCN architecture, the number of Inception modules in the Inception architecture and the number of attention heads and embedding dimensions in the ConvTran architecture. This means that all designed *student* networks are essentially quantized versions of their corresponding *teacher* networks.

### 3.4.1 Student FCN models

To evaluate the impact of convolutional filters, we constructed *student* FCN models with three layers, each having fewer convolutional filters than the *teacher* FCN model. Table 1 summarizes all the details about the designed *student* architectures, which are denoted as Model_#F in the table.

### 3.4.2 Student inception models

To assess the impact of convolutional layers, we construct *student* Inception models with fewer inception modules. Each designed *student* model has the same number of convolutional filters as the *teacher* Inception model and is denoted as Model_#M. Table 2 summarizes detailed information on all *student* Inception model configurations.

### 3.4.3 Student ConvTran models

To assess the impact of attention heads and embedding dimensions, we construct *student* ConvTran models by simultaneously reducing both components. Each designed *student* model is denoted as Model_#H. Table 3 summarizes detailed information on all *student* ConvTran model configurations.

**Table 2** Configurations of *teacher* Inception and *student* Inception architectures

| | Model name | # Inception modules | Total param. # |
|---|---|---|---|
| Teacher | Model_6M | 6 | 422 627 |
| Students | Model_5M | 5 | 325 347 |
| | Model_4M | 4 | 244 963 |
| | Model_3M | 3 | 164 579 |
| | Model_2M | 2 | 83 555 |
| | Model_1M | 1 | 3 171 |

**Table 3** Configurations of *teacher* ConvTran and *student* ConvTran architectures

| | Model name | # Attention heads | Embedding dimension | Total param. # |
|---|---|---|---|---|
| Teacher | Model_8H | 8 | 16 | 33 026 |
| Students | Model_6H | 6 | 12 | 24 692 |
| | Model_4H | 4 | 8 | 16 434 |
| | Model_2H | 2 | 4 | 8 312 |
| | Model_1H | 1 | 2 | 4 277 |

# 4 Experimental evaluation

## 4.1 Experimental setup

### 4.1.1 Data

For all our experiments, we use the UCR Archive datasets to assess the impact of knowledge distillation for TSC. This archive is the largest available repository of time series datasets and contains 128 univariate time series datasets from various domains and sensors, each with distinct characteristics such as time series length and number of samples. The number of classes in these datasets ranges from 2 to 60. We adopt the original train/test split and apply z-normalization to all time series for a fair evaluation comparable to the state-of-the-art. Following standard practices for fair comparison, we discard datasets with unequal length series or missing values and the Fungi dataset, which only provides a single training case for each class label. Consequently, we consider 112 univariate time series datasets in our experiments.

### 4.1.2 Experimental protocol

In our knowledge distillation framework, we begin by training the *teacher* models (*teacher* FCN, *teacher* Inception and *teacher* ConvTran) five times for each dataset. From these results, we select the best *teacher* model based on the lowest training loss value, ensuring reproducibility in practical applications. This best *teacher* model is then used to guide the training of the *student* model as described in Sect. 3. We propose different *student* models by reducing the number of convolutional filters, Inception modules and attention heads with embedding dimensions from the *teacher* FCN, *teacher* Inception and *teacher* ConvTran architectures, respectively. In parallel to the distilled *student* models, we also train the same *student* models without applying knowledge distillation, referred to as *studentAlone* models. First, we compare the *student* and *studentAlone* models to assess the impact of knowledge distillation. Next, we compare the *student* models with the *teacher* model to evaluate how well the *students* perform relative to the *teacher*. To reduce dependence on random initialization, we train both *student* and *studentAlone* models five times. Finally, for each dataset, we compare performance of each model by averaging the accuracy across five runs.

#### 4.1.3 Comparison protocol and metrics

Each *student* model performance is compared first with the *studentAlone* model and then with the *teacher* model across 112 UCR Archive datasets. For that reason, we take into account the number of datasets where a specific model wins, i.e., performs better than another model in terms of classification accuracy. Additionally, we account for the number of datasets in which performance results in ties and losses.

### 4.2 Hyperparameters in knowledge distillation

Knowledge distillation involves two hyperparameters, namely the distillation weight $\lambda$ and the temperature $\tau$. As shown in Eq. 6, $\lambda$ controls the trade-off between the supervised student loss and the distillation loss while $\tau$ softens the output distributions of both *teacher* and *student* in the KL term. We set $\lambda = 0.5$ to assign equal weight to both the student loss and the distillation loss and $\tau = 10$ following the standard setting widely adopted in the KD literature [17]. By using fixed hyperparameters across all datasets and architectures, we avoid the risk of overfitting to the test set through hyperparameter selection.

#### 4.2.1 Implementation details

The *teacher* FCN, Inception and ConvTran models were implemented following the original configurations described in [41], [21] and [12], respectively. For each architecture, we separately analyze impact of knowledge distillation by reducing the complexity of *student* models. PyTorch is the main framework in our implementation which is used to build all *teacher*, *student* and *studentAlone* models. All models are trained using the Adam optimization algorithm with a batch size of 16, 64 and 64 for the FCN, Inception and ConvTran models, respectively. Additionally, we utilize a learning rate scheduling mechanism that dynamically reduces the learning rate based on performance with a reduction factor of 0.5, a patience value of 50 epochs and a minimum learning rate of 0.001. All experiments were conducted on a system equipped with an NVIDIA RTX 5090 GPU (32GB memory), running Ubuntu 22.04, with Python 3.12 and PyTorch 2.5.1. Our code is publicly available at https://github.com/MSD-IRIMAS/KD-4-TSC.

### 4.3 Experimental results

#### 4.3.1 Performance comparison of student FCN models

To evaluate the impact of convolutional filters in our KD framework, we analyze their effect on the performance of *student* FCN models by comparing *student* models with *studentAlone* models. As illustrated in Fig. 4a, we assess the number of wins for each model. When the number of filters is high, the performance of *studentAlone* models surpasses that of *distilled student* models. With an intermediate number of convolutional filters, the *distilled student* models effectively leverage the knowledge of the *teacher* model, achieving superior performance compared to *studentAlone* models. However, when the number of filters is low, knowledge distillation results in suboptimal performance as the *distilled student* models deteriorate. These findings provide deeper insights into how the number of convolutional filters influences the performance of different *student* models within the KD framework.

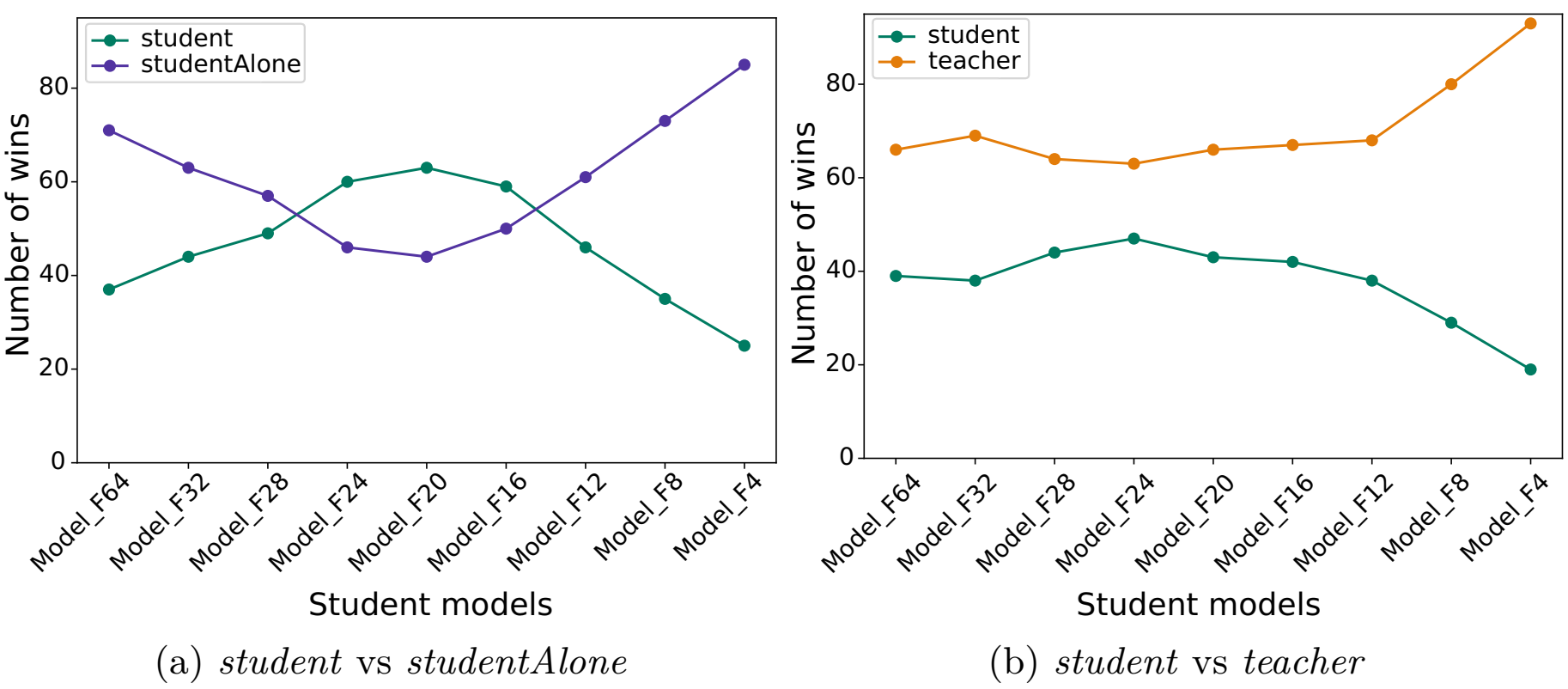

**Fig. 4** Number of wins considering classification accuracies according to various numbers of student FCN models filters

Hence, the most effective *student* FCN configuration utilizing KD is the architecture that consists of 20 filters in the first and third layers and 40 filters in the second layer. For this specific configuration, Fig. 5 presents the accuracy plot of the *student* model compared to the *studentAlone* model across the 112 UCR datasets.

An interesting observation from Fig. 5 is the asymmetry in win magnitudes. When the *student* model wins, improvements are typically modest (points near the diagonal), whereas *studentAlone* achieves a few substantial wins (points far from the diagonal). This asymmetry occurs because KD acts as a regularization mechanism that prevents extreme failures but also limits exceptional gains by guiding the *student* toward performance level of the *teacher*. In contrast, the *studentAlone* can achieve larger improvements on datasets where the architecture is naturally well-suited. Additionally, our fixed hyperparameters may not be optimal for all datasets in diverse UCR Archive which potentially limits the effectiveness of KD in such cases.

Next, we compare the performance of the distilled *student* models with the *teacher* model. The number of wins for each model is illustrated in Fig. 4b. These results reveal trends similar to those observed earlier. The distilled *student* models with an intermediate number of filters exhibit competitive performance with the *teacher* model. Moreover, the *student_8F* model surpasses the *teacher* model in approximately one-third of the datasets, despite having fewer than 1% of the *teacher* model parameters.

### 4.3.2 Performance comparison of student inception models

In this section, we analyze the impact of inception modules on the *student* Inception models. For this purpose, we primarily focus on comparing the performance of the distilled *student* models against the *studentAlone* models. The number of wins for each model is depicted in Fig. 6a. The results show that when the number of Inception modules is high, in this case 5, KD slightly degrades the performance of the distilled *student_5M* model. Reducing the number of inception modules to 4 significantly boosts the performance of the distilled *student_4M* model through KD. Finally, for the relatively smaller number of inception modules, *student_3M*, *student_2M* and *student_1M* models, KD degrades the performance of the distilled *student* models.

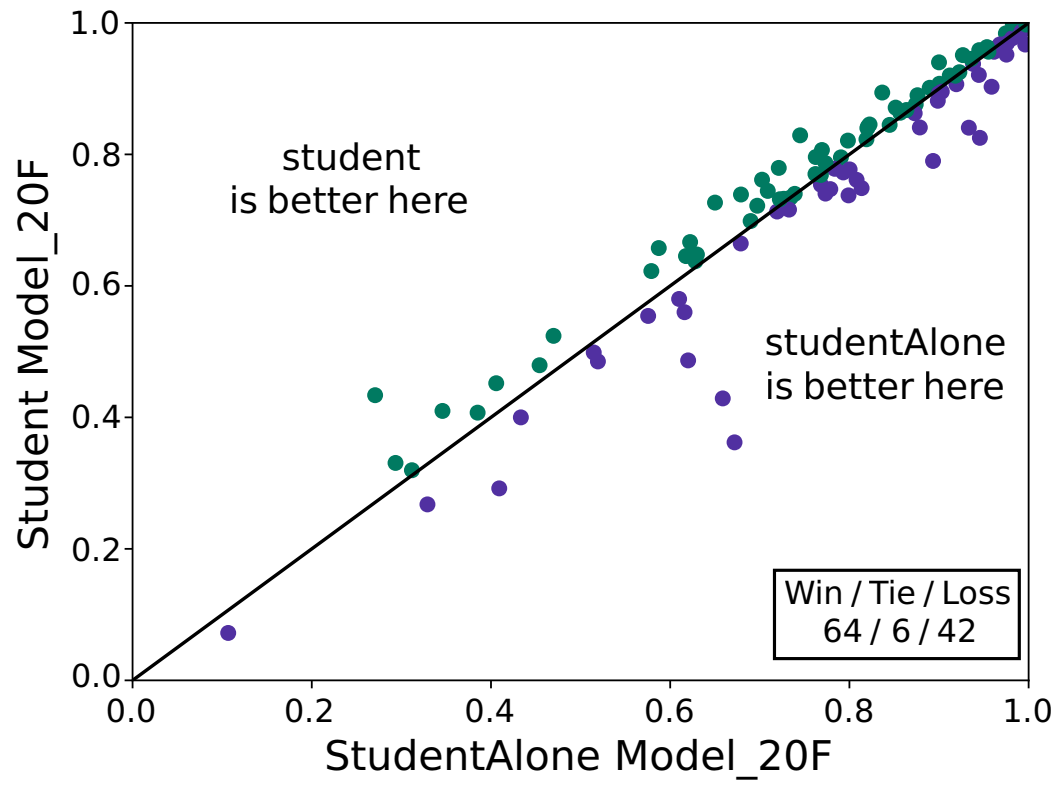

**Fig. 5** Accuracy plot shows performance of student_20F model against to studentAlone_20F FCN model

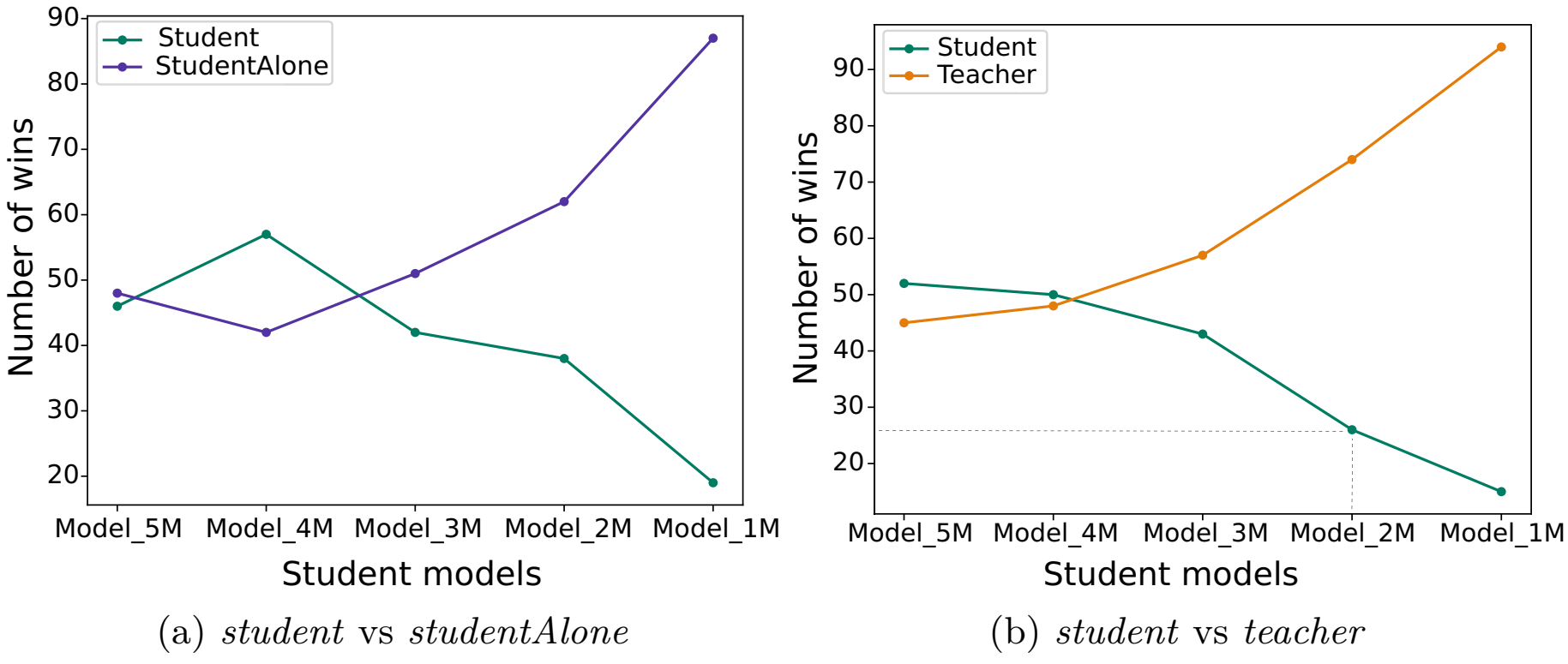

**Fig. 6** Number of wins considering classification accuracy according to various numbers of Inception modules of student models

Hence, the best distilled *student* Inception network leveraging KD is the *student_4M* Inception architecture which has 16 more wins over the *studentAlone_4M* Inception model. The *student_4M* Inception model has two more wins over the *teacher* Inception model. Considering that the *student_4M* Inception model contains 42% fewer parameters than the *teacher* Inception model, it can be regarded as a very promising architecture to replace the *teacher* Inception model. Figure 7, depicts the accuracy plot of the *student_4M* Inception model against the *teacher* Inception model for each of the 112 UCR archive datasets. From the figure, we can notice that the *student_4M* model nearly achieves the same performance as the *teacher* model while requiring less training and inference time.

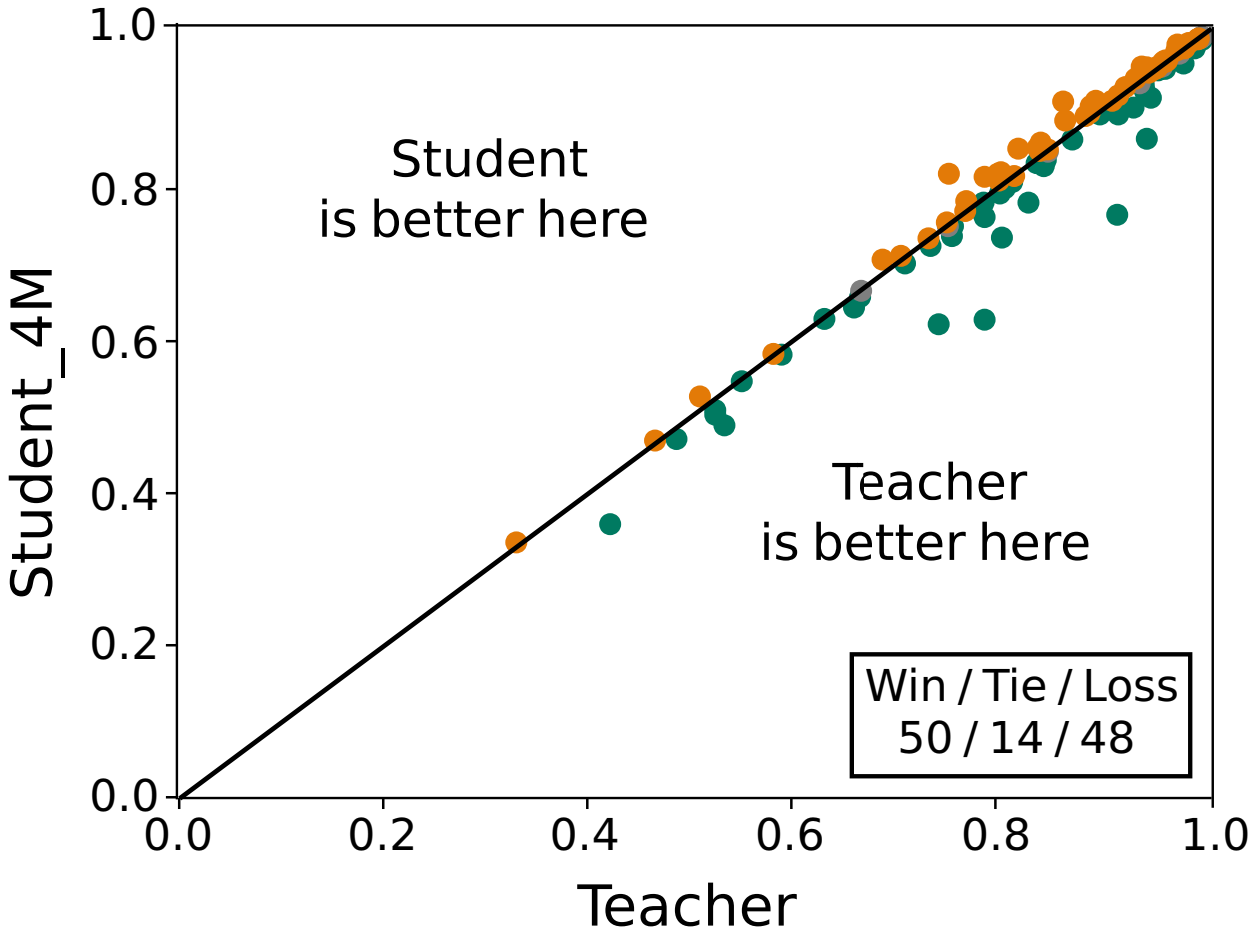

**Fig. 7** Accuracy plot shows performance of the *student_4M* with 42% less parameters against to the *teacher*

Lastly, we analyze the performance of the *student* Inception models in comparison with the *teacher* Inception model. Figure 6b presents the number of wins for each model. The results indicate that the *student_5M* and *student_4M* models achieve performance comparable to the *teacher* model. Additionally, the *teacher* model significantly outperforms the *student* models with one or two Inception modules. However, as shown in Fig. 6b

#### 4.3.3 Performance comparison of student ConvTran models

In this section, we analyze the impact of reducing the number of attention heads and embedding dimensions on the *student* ConvTran models. We first compare the performance of the distilled *student* models against the *studentAlone* models. The number of wins for each model is depicted in Fig. 8a. The results show that KD improves the performance of the distilled *student* ConvTran models for smaller configurations. The most significant improvement is observed for the *student_2H* model which achieves 62 wins against 42 wins for the *studentAlone_2H* model. As the number of attention heads and embedding dimension increase, the gap between *student* and *studentAlone* models narrows, with both achieving nearly equal performance at 6 attention heads.

Hence, the best distilled *student* ConvTran model leveraging KD is the *student_2H* architecture with 2 attention heads and embedding dimension 4, which has 20 more wins over the *studentAlone_2H* model. For this specific configuration, Fig. 9 presents the accuracy plot of the *student_2H* model compared to the *studentAlone_2H* model across the 112 UCR datasets. The *student_2H* model wins on 62 datasets against the *studentAlone_2H* model, confirming that KD provides a substantial performance improvement for the ConvTran architecture at this complexity level.

Next, we compare the performance of the distilled *student* models with the *teacher* model. The number of wins for each model is illustrated in Fig. 8b. The results indicate that as the number of attention heads and embedding dimension increase, the *student* models achieve increasingly competitive performance against the *teacher* model. The *student_6H* model surpasses the *teacher* model on 57 out of 112 datasets. However, as shown in Fig. 8a, the gap between the *student_6H* and *studentAlone_6H* models is negligible, indicating that KD provides minimal additional benefit at this complexity level. This is consistent with the behavior observed in both FCN and Inception architectures, where student models close to the teacher in terms of model size do not significantly benefit from KD. At the other end, the *student_1H* model fails to compete with the *teacher* model, with the *teacher* winning on 105 out of 112 datasets. Therefore, the most meaningful contribution of KD for the ConvTran architecture is observed for intermediate student configurations, particularly the *student_2H* model where KD provides a substantial performance boost over the *studentAlone* model.

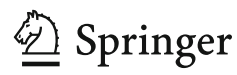

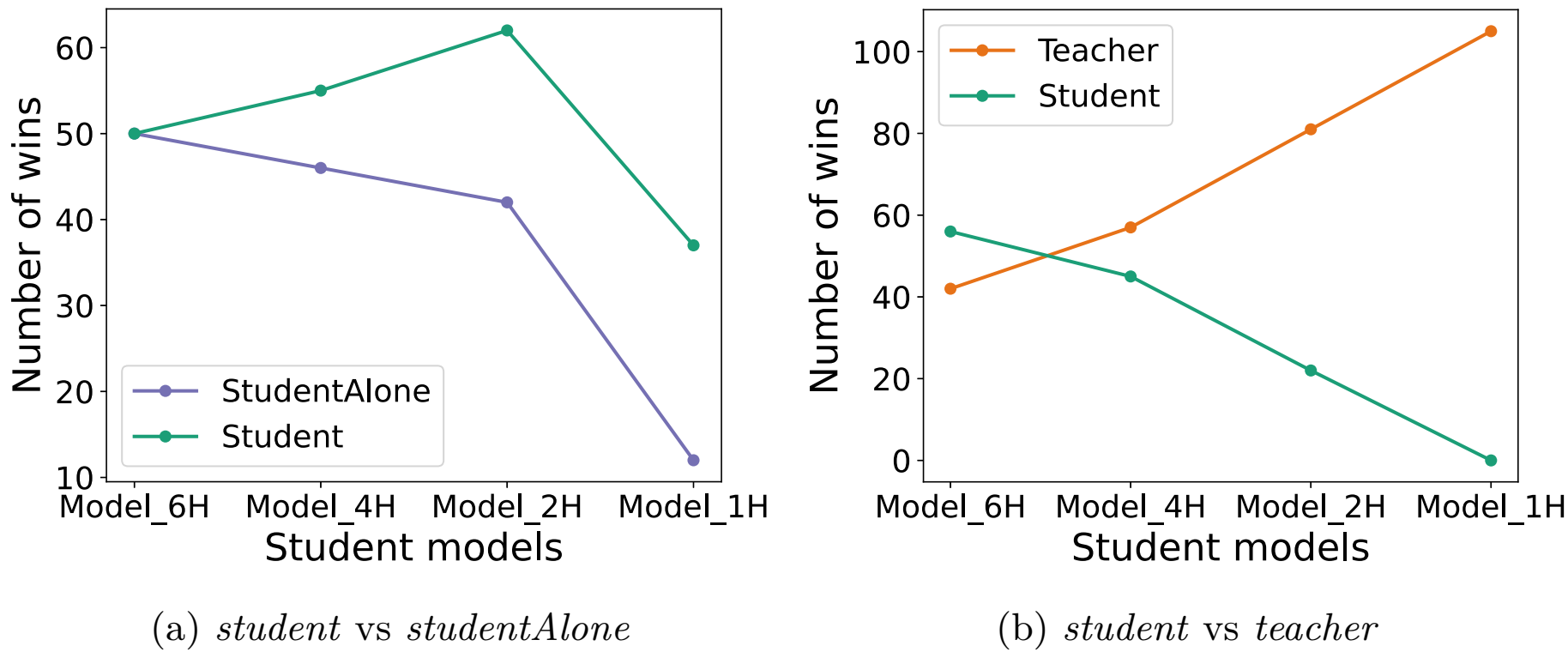

**Fig. 8** Number of wins considering classification accuracy according to various numbers of attention heads and embedding dimensions of student ConvTran models

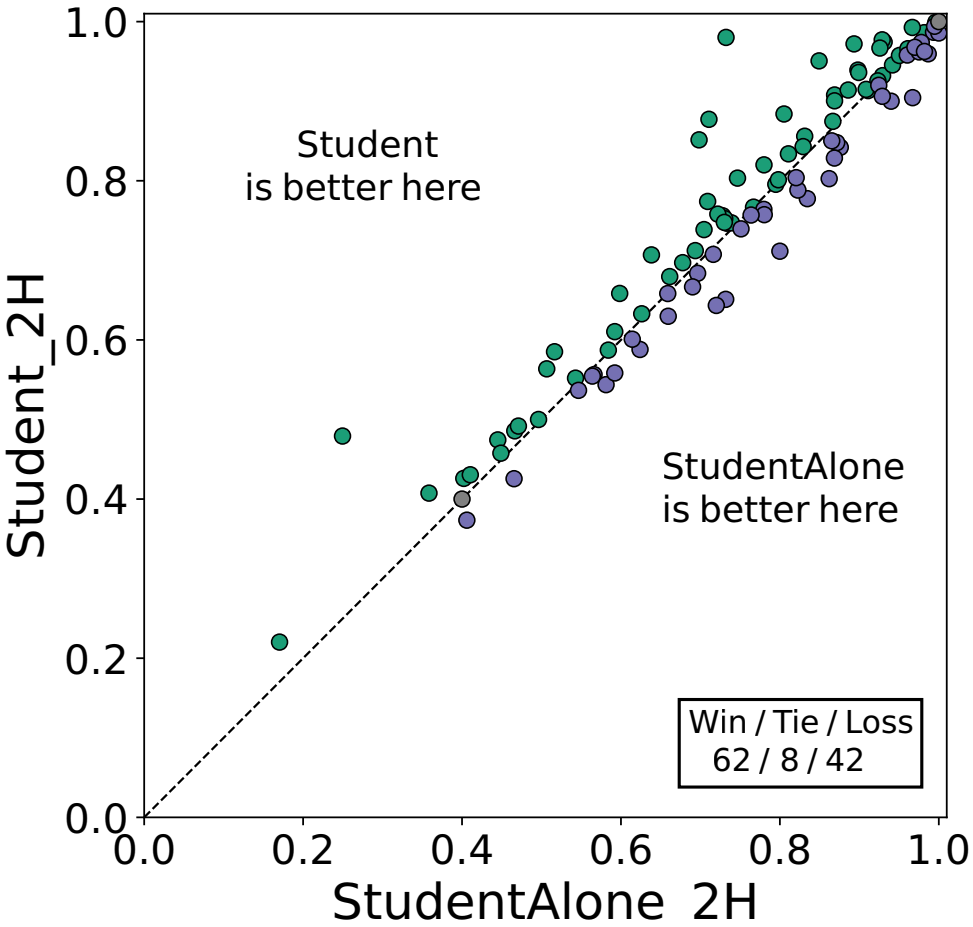

**Fig. 9** Accuracy plot shows performance of the *student_2H* model against the *studentAlone_2H* ConvTran model

### 4.4 Experimental analysis

#### 4.4.1 Relationship between KD and model complexity

Across all three architectures, in the case of more complex *student* models, KD degrades the performance of the distilled *student* models. This behavior can be explained by the fact that complex *student* models are close to the *teacher* model in terms of model size, allowing them to capture distinctive features independently and achieve highly competitive results with the *teacher* model. Therefore, forcing the complex distilled *student* models to mimic the *teacher* model prevents them from learning novel representations that could improve performance. Next, in the case of intermediate complexity *student* models, KD helps to boost the performance of the distilled *student* models across all three architectures. This can be explained by the fact that the complexity of these *student* models is large enough to capture and leverage knowledge from a complex *teacher* model. Hence, KD enhances their ability to generalize better by learning from the *teacher* distilled knowledge. Finally, in the case of less complex *student* models, KD degrades the performance of the distilled *student* models for both FCN and Inception architectures. As shown in [29], a smaller *student* network may lack the capacity to utilize the knowledge of a larger *teacher* network which may explain the performance degradation in the smaller distilled *student* models. However, for the ConvTran architecture, KD continues to improve the performance of even the smallest *student_1H* model, suggesting that the self-attention mechanism may be more receptive to knowledge transfer than purely convolutional architectures at lower model capacities.

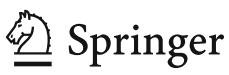

From Figs. 4b and 6b, we observe that smaller student models achieve a considerable number of wins over their respective teacher models across the UCR Archive datasets. Specifically, the smaller student FCN models (*student_12F* and *student_8F*) outperform the teacher FCN model on more than 28 out of 112 datasets while the *student_2M* Inception model surpasses the teacher Inception model on 27 out of 112 datasets. These results demonstrate that smaller student models, except for very small configurations, achieve higher accuracy than the teacher model on roughly one-fourth of the UCR Archive datasets. This can be explained by the fact that smaller *student* models have significantly smaller receptive fields which makes them less prone to overfitting on smaller datasets. From another point of view, this also shows large variability in the UCR Archive where it is not possible to identify a particular architecture with a certain number of layers and filters that performs well on all datasets.

### 4.4.2 Relationship between KD and filter transfer

This subsection aims to visually analyze convolutional filters of the *teacher* Inception, *student* Inception and *studentAlone* Inception models to comprehend the real impact of KD on the Inception network. As we mentioned, the original Inception model consists of six Inception modules, each containing four convolutional filters of sizes 40, 20, 10 and 1. In our experiments, we visualize the convolutional filters from the first inception module for the EOGVerticalSignal dataset. From the first inception module, we selected the filter with size 40 because it is the longest filter that can represent learned parameters in a better way than the others. In our experiments, we use the default settings from the original *InceptionTime* paper [21], so the number of filters is 32. Consequently, the convolutional filter shape for the model is 32 x 40. We use the DTW distance measure to compute the distance between each pair of filters and then, by utilizing the T-distributed Stochastic Neighbor Embedding (T-SNE) technique, we project them into a two-dimensional coordinate system. The T-SNE technique transforms the filters into a 32 x 2 matrix which can be easily visualized in a Cartesian coordinate system.

The *student_4M*, *studentAlone_4M*, and *teacher* Inception models achieved accuracy scores of 0.483, 0.447 and 0.477, respectively, on the EOGVerticalSignal dataset. Since the performance of the *student* model is closer to that of the *teacher* model, we initially expected its convolutional filter distribution to more closely resemble that of the *teacher* model rather than the *studentAlone* model. To explore this, we first compare the convolutional filters of the *teacher* and *studentAlone* models. Figure 10a visualizes the convolutional filters from the first Inception module of both models, where purple and orange points represent the *studentAlone* and *teacher* models, respectively. The plot reveals that the distribution of *studentAlone* convolutional filters (purple points) generally differs from that of the *teacher* filters (orange points). For instance, the purple circle highlights a region where the *studentAlone* filters exhibit a distinct pattern compared to the *teacher* filters. Additionally, within the orange dotted line, a small number of purple points indicate minimal overlap. This variation in distribution suggests that the *teacher* and *studentAlone* models generalize differently, which may explain the observed performance differences despite our initial expectations.

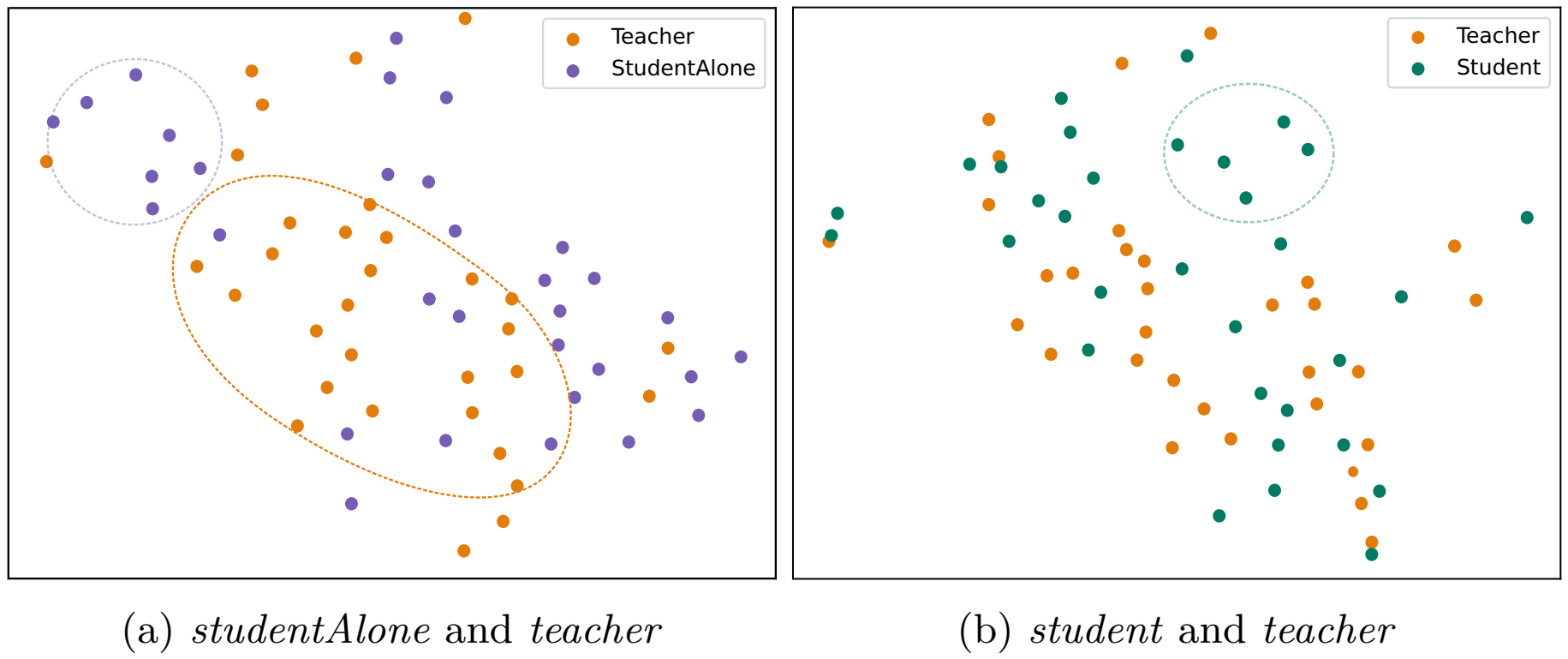

(a) *studentAlone* and *teacher* (b) *student* and *teacher*

**Fig. 10** Visualization of trained filters on EOGVerticalSignal dataset

Next, we conducted the same experiments for the *teacher* and *student* models. Figure 10b visualizes the convolutional filters of both models, where orange and green points represent the *teacher* and *student* models, respectively. The figure shows that these points are distributed more compactly, indicating a greater similarity between the *student* and *teacher* filters compared to the *studentAlone* filters. Additionally, the small green circle highlights a few *student* filters that exhibit different patterns from those in the *teacher* model, suggesting that while the *student* model closely follows the *teacher*, it still captures some distinct features.

The aim of KD is to let the *student* model mimic the predictions of the *teacher* model. As mentioned in Section 3, during the training process, the *teacher* model also transfers its generalization ability to the *student* model. From Figs. 10a and 10b, we can conclude that the *teacher's* convolutional filters are more similarly distributed to the *student* filters than the *studentAlone* filters, demonstrating the effectiveness of KD.

### 4.4.3 Relationship between KD and model generalization

From the previous experimental results, we discovered that the *student_4M* Inception model benefited the most from KD and demonstrated the highest performance improvement over the *studentAlone_4M* Inception model. Since the *student_4M* model leverages knowledge from the pre-trained *teacher* Inception model, we hypothesize that it possesses superior generalization ability on unseen time series data.

To validate this hypothesis, we compare the training and validation loss curves for *student_4M* and *studentAlone_4M* models on the EOGVerticalSignal dataset, as shown in

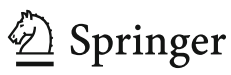

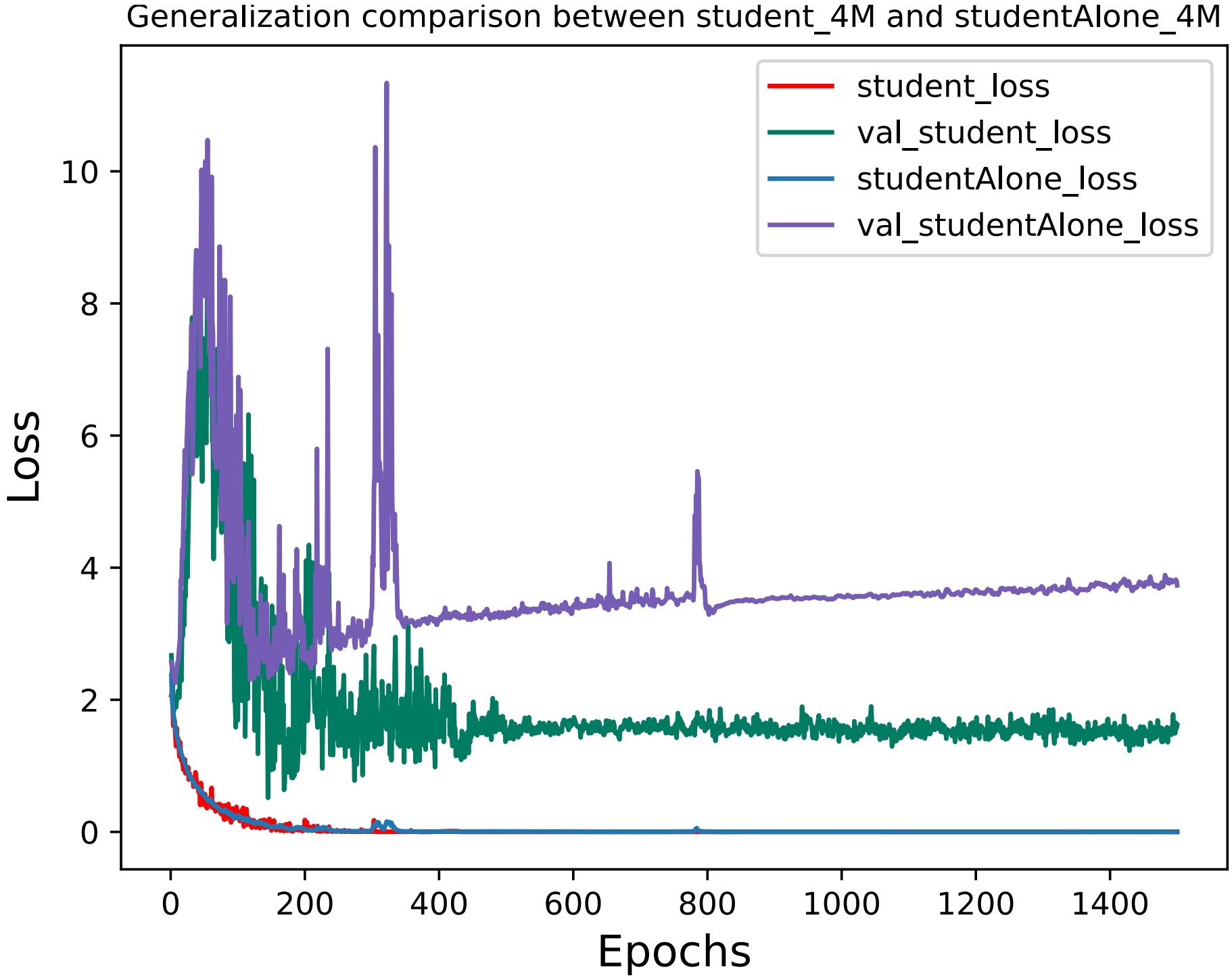

**Fig. 11** Generalization comparison between student_4M and studentAlone_4M

Fig. 11. To ensure a fair comparison, we consider only classification loss for the *student_4M* model, excluding the distillation loss. Since the UCR archive does not provide separate validation data, we use test set to monitor generalization behavior during training. Thus, the validation loss is computed based on the test data.

Figure 11 reveals two key observations. First, the validation loss of the *student_4M* model converges to a considerably lower value than that of the *studentAlone_4M* model. Second, the generalization gap of the *student_4M* is approximately half that of the *studentAlone_4M* model. Notably, the *studentAlone_4M* model exhibits clear overfitting behavior. While the training loss of the *studentAlone_4M* model continues to decrease, the validation loss plateaus and slightly increases after approximately 200 epochs. In contrast, the *student_4M* model shows consistent convergence of both training and validation losses without overfitting. These results suggest that knowledge distillation acts as an effective regularization mechanism. The soft targets from teacher model encode richer information about inter-class relationships and prediction uncertainty, helping the student model learn more generalizable representations. This improved generalization directly explains the 3.6% higher test accuracy of the *student_4M* model compared to the *studentAlone_4M* model on the EOGVerticalSignal dataset.

## 5 Conclusion and future works

In this paper, we examined the impact of KD on three architectures by reducing the number of convolutional filters (FCN), Inception modules (Inception) and attention heads with embedding dimensions (ConvTran). The experiments were conducted on 112 datasets from the UCR Archive 2018, the largest repository of time series data available. Our results show that KD significantly improves performance of *student* models with intermediate complexity across all three architectures, consistent with findings in [29] for image classification. For FCN, the *student_20F* model reduces parameters by a factor of 38 while maintaining competitive performance. For Inception, the *student_4M* model achieves two more wins than the *teacher* with 42% fewer parameters, standing out as a highly efficient alternative. For ConvTran, the *student_2H* model benefits the most from KD with 20 more wins than the *studentAlone_2H* model. Notably, KD continues to improve ConvTran performance even for the smallest student configurations, unlike FCN and Inception where very small students fail to benefit from distillation. We also observed that distilled student convolutional filters are more similar to teacher filters than those of the studentAlone model, and that KD leads to better generalization. Although KD is typically evaluated on a few datasets in image classification, our evaluation across 112 datasets shows that finding an appropriate architecture for all datasets is not straightforward. Our study focuses on establishing a large-scale baseline for *response-based* KD on UCR benchmarks. As future work, we plan to investigate time series-aware distillation strategies such as alignment-based (soft-DTW-style) losses between teacher and student *temporal feature sequences*.

**Acknowledgements** This work was supported by the ANR DELEGATION project (grant ANR-21-CE23-0014) of the French Agence Nationale de la Recherche. The authors would like to acknowledge the High Performance Computing Center of the University of Strasbourg for supporting this work by providing scientific support and access to computing resources. Part of the computing resources were funded by the Equipex Equip@Meso project (Programme Investissements d'Avenir) and the CPER Alsacalcul/Big Data. The authors would also like to thank the creators and providers of the UCR Archive.

**Author Contributions** J.A. conceptualized the study, designed the methodology, and conducted the experiments. M.D. contributed to the experimental design, data analysis, and interpretation of results. M.D., G.F. and J.W. supervised the research, provided critical feedback, and contributed to manuscript refinement. J.A. wrote the main manuscript, with M.D., G.F. and J.W. assisting in revisions and additional sections. All authors reviewed and approved the final version of the manuscript.

**Funding** Open access funding provided by Université de Haute-Alsace.

**Data Availability** For our experiments we have used open source UCR time series classification archive which is the largest time series classification repository. The link for the data: https://www.timeseriesclassification.com/

## Declarations

**Conflict of interest** The authors declare no conflict of interest.

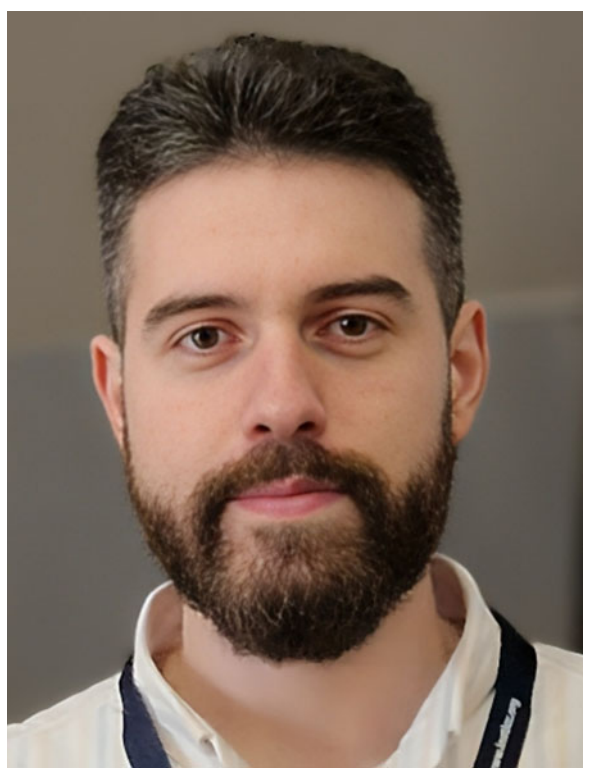

**Javidan Abdullayev** is a Ph.D. candidate in Computer Science at the University of Haute-Alsace, France, affiliated with the IRIMAS institute. His research focuses on deep learning for time-series analysis and classification with particular interest in model compression and domain adaptation. He has worked on improving the efficiency and robustness of neural network models for time-series classification tasks.

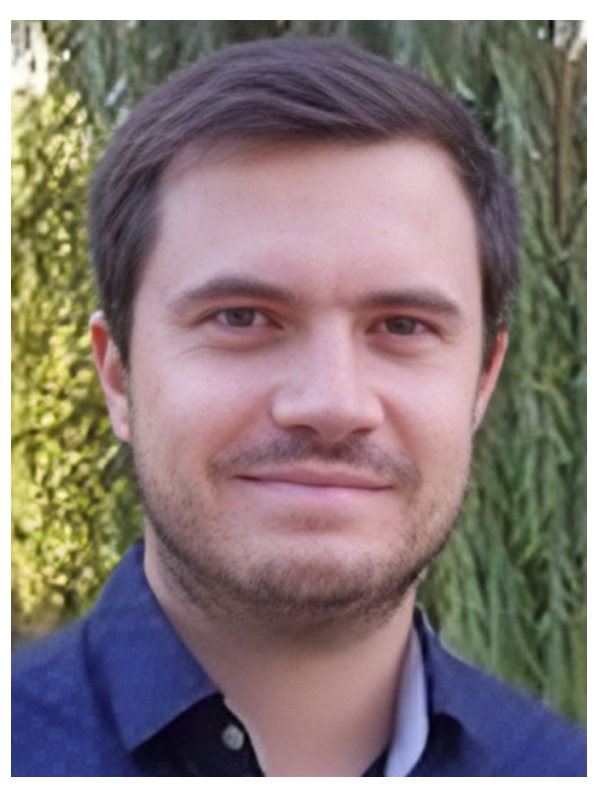

**Maxime Devanne** is Associate Professor (with habilitation) at Université Haute-Alsace in the IRIMAS Lab. His main research interests are computer vision, data mining and deep learning, with applications in time-series classification, human behavior and motion understanding.

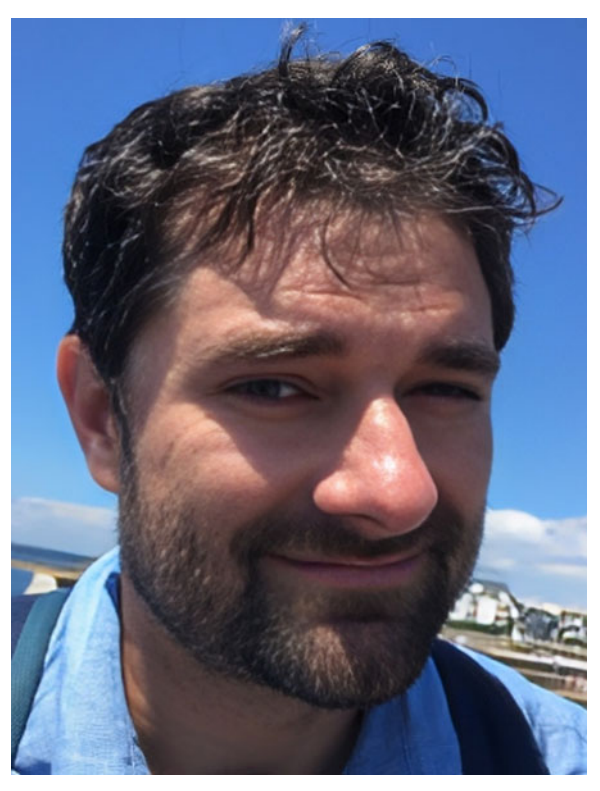

**Jonathan Weber** obtained his PhD in Computer Science in 2011 at University of Strasbourg and became assistant professor at Université de Lorraine in 2012; he moved to Université de Haute-Alsace as associate professor in 2016. Since 2023, he is full professor at Université Haute-Alsace in the IRIMAS Lab. His main research interests are computer vision, data mining, time-series classification and deep learning, with various application fields.

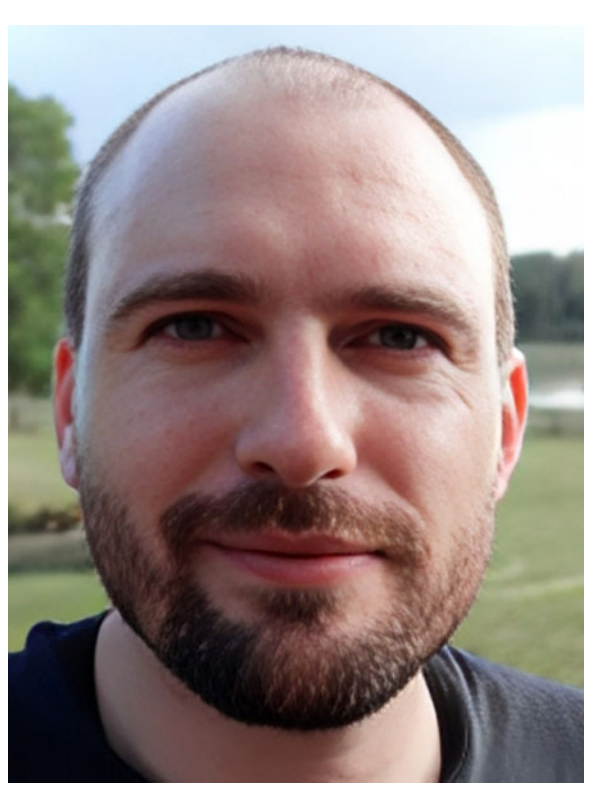

**Germain Forestier** received his PhD in Computer Science from the University of Strasbourg in 2010, followed by a postdoctoral fellowship at INRIA Rennes and INSERM. He joined the University of Haute-Alsace as an Associate Professor in 2011 and has been a Full Professor since 2018. He is also an Adjunct Professor at Monash University (Australia). His research focuses on data science, data mining, time-series analysis, machine learning, big data, artificial intelligence and deep learning.